\documentclass[10pt,twocolumn,letterpaper]{article}

\usepackage{cvpr}
\usepackage{times}
\usepackage{epsfig}
\usepackage{graphicx}
\usepackage{amsmath}
\usepackage{amssymb}

\usepackage[breaklinks=true,bookmarks=false]{hyperref}

\cvprfinalcopy 

\def\cvprPaperID{****}

\begin{document}

\title{Moving Object Proposals with Deep Learned Optical Flow for Video Object Segmentation}

\author{Ge Shi\\
Umass Amherst\\
140 Governors Dr\\
{\tt\small geshi@umass.cs.edu}
\and
Zhili Yang\\
Umass Amherst\\
140 Governors Dr\\
{\tt\small zhiliyang@umass.cs.edu}
}

\maketitle

\begin{abstract}

Dynamic scene understanding is one of the most conspicuous field of interest among computer vision community. In order to enhance dynamic scene understanding, pixel-wise segmentation with neural networks is widely accepted. The latest researches on pixel-wise segmentation combined semantic and motion information and produced good performance. In this work, we propose a state of art architecture of neural networks to accurately and efficiently get the moving object proposals (MOP~\cite{o2015learning}). We first train an unsupervised convolutional neural network (UnFlow~\cite{meister2018unflow}) to generate optical flow estimation. Then we render the output of optical flow net to a fully convolutional SegNet model. The main contribution of our work is (1) Fine-tuning the pretrained optical flow model on the brand new DAVIS Dataset \cite{pont20172017}; (2) Leveraging fully convolutional neural networks with Encoder-Decoder architecture to segment objects. We developed the codes with TensorFlow, and executed the training and evaluation processes on an AWS EC2 instance.
   
\end{abstract}

\section{Introduction}

The Convolutional Neural Networks (CNN~\cite{o2015introduction}) has achieved great success in learning image features. Through Encoder-Decoder model, the convolutional neural networks can preserve spatial information when extracting general feature maps. This model greatly promoted the performance of convolution neural networks when handling pixel-wise segmentation \cite{badrinarayanan2017segnet} and object detection \cite{redmon2016you} tasks. Pixel-wise segmentation \cite{fayyaz2016stfcn} is to assign each pixel an class based on its semantic meaning or which object it belongs to.

\begin{figure}[ht]
\begin{center}
   \includegraphics[width=1.0\linewidth]{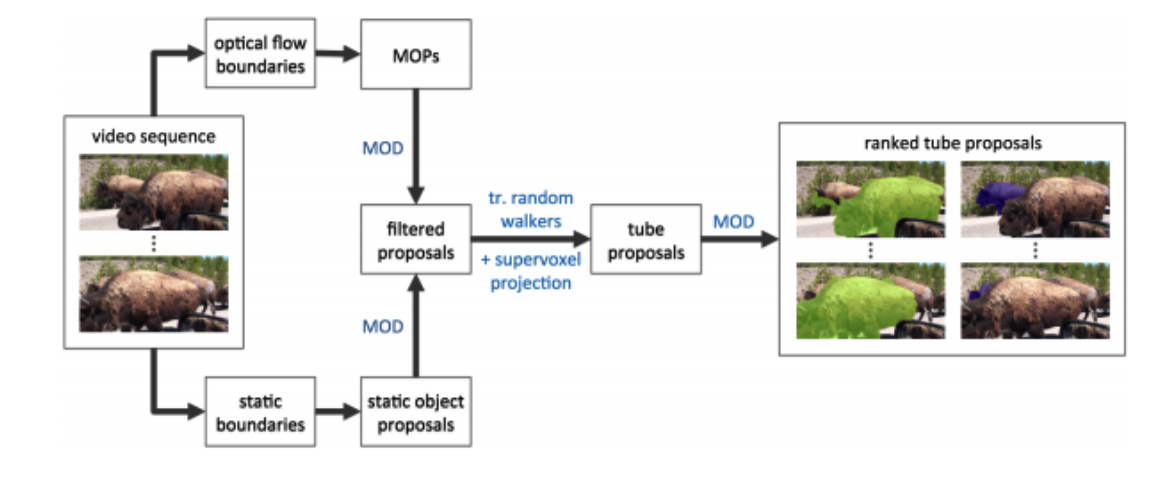}
\end{center}
   \caption{Fusion static object information and motion information to segment images.}
\label{fig:fusion}
\end{figure}

\begin{figure*}[ht]
\begin{center}
	\includegraphics[width=0.8\linewidth]{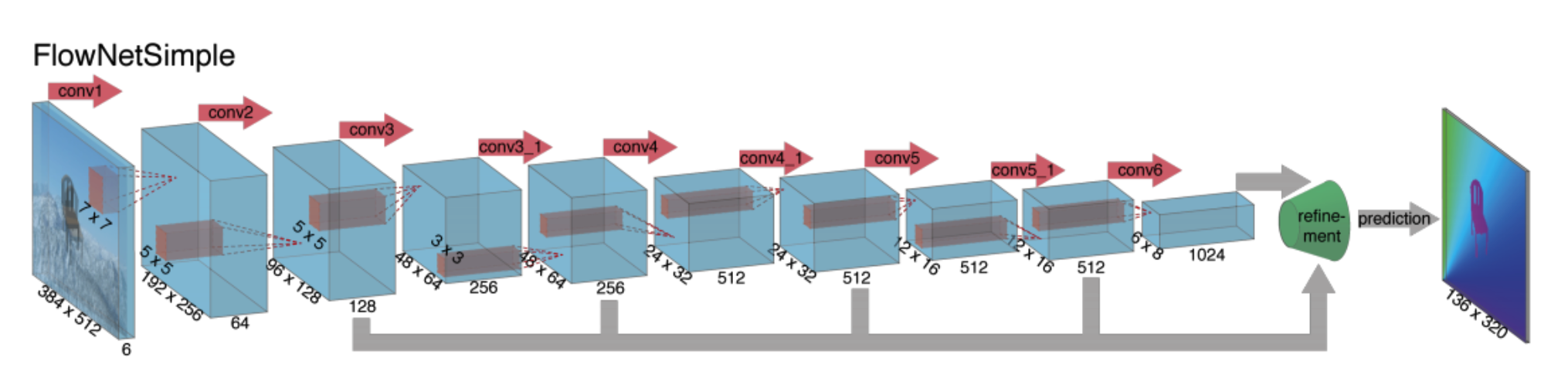}
    \includegraphics[width=0.8\linewidth]{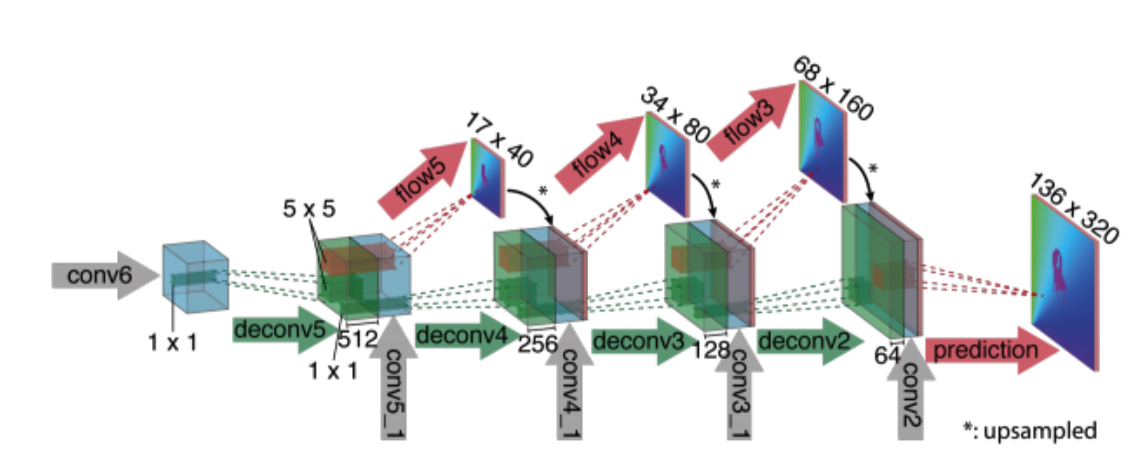}
\end{center}
   \caption{Convolution and deconvolution layers}
\label{fig:conv-dconv}
\end{figure*}

Video object segmentation \cite{dutt2017fusionseg, haque2017joint} is the task of extracting spatial regions corresponding to objects from a sequence of image frames (Figure~\ref{fig:fusion}). This is not just a process to apply image segmentation algorithm to all frames in the video, but the variational information with respect to time base also facilitates better understanding of object independence. The traditional methods solving this problem \cite{zhang1996survey, cheng2001color} relies on hand-crafted features without leveraging CNN to learn video representation. Recently, proposal of regions that likely contains independent objects and using CNN currently dominate object detection in static images. Object segmentation \cite{shi2015convolutional} is on the basis of proposed regions of interest object. When it comes to segmentation of moving objects in videos, \cite{fragkiadaki2015learning} came up with a paradigm for detecting moving objects in videos by introducing moving 
object proposals (MOP). MOP~\cite{o2015learning} is a motion based proposal method to generate a set of region proposals in each frame using segmentations on optical flow \cite{sun2010secrets}. To get impressive results in objects segmentation in video, many advanced works \cite{dosovitskiy2015flownet, ilg2017flownet, ren2017unsupervised, meister2018unflow} unites optical flow boundaries coming from motion information and static boundaries coming from static object proposals.

SegNet~\cite{badrinarayanan2017segnet} is a deep fully convolutional neural network architecture for semantic pixel-wise segmentation, for which the model can be divided into two parts, encoder and decoder (Figure~\ref{fig:conv-dconv}). The encoder contains the 13 convolutional layers adapted from VGG16 \cite{simonyan2014very} network, which is commonly used in many segmentation architectures; the decoder uses pooling indices computed in the max-pooling step of the corresponding encoder to perform non-linear upsampling. In our project, instead of semantic segmentation, we adapted the source codes to achieve moving object segmentation in videos.

Optical flow is another traditional computer vision research area which estimates the relative motion of objects and edges in a visual scene. The algorithms utilizing neural networks to address the task have been invented recently. Researchers explored both innovative supervised and unsupervised structures of optical flow learning networks. FlowNet \cite{dosovitskiy2015flownet} and UnFlow \cite{meister2018unflow} are representatives of them. 

In this work, we decided to develop a neural networks to segment objects in video leveraging unsupervised learning neural networks to train optical flow and feed the outputs of optical flow estimation to into a SegNet model.

We choose DAVIS 2017 \cite{pont20172017} as the dataset for training and evaluation, which contains 50 high quality video sequences and so as the corresponding ground truth for evaluation. Each video of the dataset is not limited in one specific object, instead it might contain multiple objects in one frame and each of them are annotated in the ground truth.

\section{Related Works}

\subsection{Methods for Estimating Optical Flow}
\noindent
\textbf{FlowNet \cite{dosovitskiy2015flownet}:} End-to-end supervised learning of convolutional networks for optical flow was first introduced with FlowNet. The network takes two consecutive input images and outputs a dense optical flow map using an encoder-decoder architecture. For supervised training, a large synthetic dataset was generated from static background images and renderings of animated 3D chairs. A follow up work \cite{ilg2017flownet} introduced the more accurate, but also slower and more complex FlowNet2 family of supervised networks.

\noindent
\textbf{UnFlow:} \cite{meister2018unflow} built on the previous FlowNet-based networks and extend it unsupervised video segmentation settings. First, they designed a symmetric, occlusion-aware loss based on bidirectional (i.e., forward and backward) optical flow estimates. Second, they trained FlowNetC with their comprehensive unsupervised loss to estimate bidirectional flow. Third, they made use of iterative refinement by stacking multiple FlowNet networks. Optionally, they can also use a supervised loss for fine-tuning their networks on sparse ground truth data after unsupervised training.

\subsection{Methods for Object Segmentation}
\noindent
\textbf{PCM \cite{bideau2016s}:} In this work, a likelihood function uses a novel combination of the angle and magnitude of the optical flow to maximize the information about the truth motions of objects is derived. It is used for assessing the probability of an optical flow vector given the 3D motion direction.

\noindent
\textbf{MP-Net \cite{tokmakov2017pattern}:} MP-Net learns to recognize motion patterns in a flow field. Despite its decent performance, it is limited by its frame-based nature and also overlooks appearance features of objects.

\noindent
\textbf{ConvGRU-Net \cite{tokmakov2017learning}:} ConvGRU-Net is a novel approach for video object segmentation. It combines two complementary sources of information: appearance and motion, with a visual memory module, realized as a bidirectional convolutional gated recurrent unit. The ConvGRU module encodes spatio-temporal evolution of objects in a video and uses this encoding to improve motion segmentation.

\section{Technical Approach}

Our method to segment moving objects can be separated as two steps: (1) training an unsupervised neural networks to learn optical flow; (2) feeding a SegNet model with the learned output from optical flow networks and generate moving object segmentation. The details are as following.

\subsection{Optical Flow}

Since suggested \cite{horn1981determining}, most dense approaches estimating optical flow are based on the variational formulation. According to the optical flow constraint, the gray value of a moving pixel stays constant over time, i.e.

\begin{figure}[t]
\begin{center}
   \includegraphics[width=1.0\linewidth]{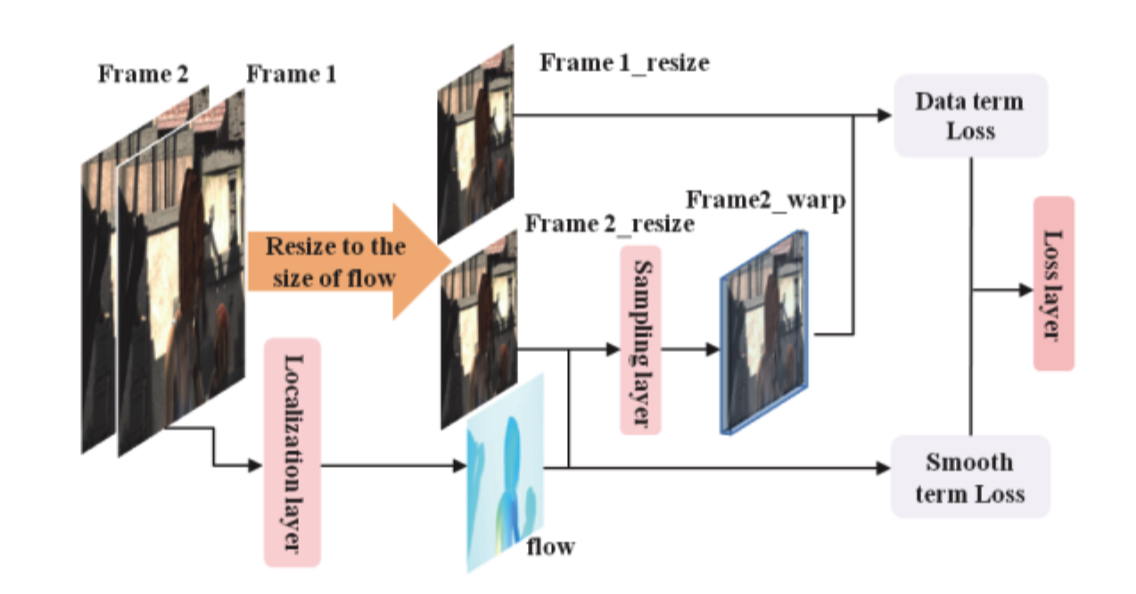}
\end{center}
   \caption{Unsupervised learning model based on variational constraint.}
\label{fig:optical}
\end{figure}

$$
I(x,y,t) = I(x+u, y+v, t+1)
$$
where $(u,v)$ is the optical flow vector of a pixel $(x,y)$ from time t to time t +1.
Using the first-order Taylor approximation, the linearized version of the image function can be derived.
$$
I (x,y,t) \approx I (x,y,t+1) + \nabla I (x,y,t+1)^T (u,v)
$$
$$
0 = I (x,y,t+1) - I (x,y,t) + \nabla I (x,y,t+1)^T (u,v)
$$
The partial derivatives of the image function can be denoted as $I_t$ , $I_x$ ,and $I_y$. The optical flow constraint equation has inherent dependency on the image position (x,y).
$$
0 = I_t + I_x u + I_y v.
$$
The smoothness penalty which penalizes the derivative of the optical flow field, was introduced by the variational methods (Figure~\ref{fig:optical}),
$$
\min_{u(x),v(x)} \{  \int_{\Omega} (| \nabla u(x,y) |^2 +| \nabla v(x,y) |^2) d \Omega +
$$
$$
\lambda \int_{\Omega} (I_t + I_x u(x,y) + I_y v(x,y)) d \Omega \}
$$
where $\Omega$ is the image domain.
The improvement of basic variational energy function has been exhaustively studied by researchers. We adopt the the Charbonnier penalty $\Psi(x) = \sqrt {(x^2 + \epsilon^2)}$ and here $\epsilon$ is 0.001. Then we have two terms of loss function. 
$$
l_d = \lambda \int_{\Omega} \Psi (I_1(x,y) - I_2(x+u(x,y),y+v(x,y))) + 
$$
$$
\Psi (\nabla I_x) + \Psi (\nabla I_y) d\Omega
$$
$$
l_s = \int_{\Omega} \Psi (u(x,y) - u(x+1,y)) + \Psi (u(x,y+1) - u(x,y)) + 
$$
$$
\Psi (v(x,y) - v(x+1,y)) + \Psi (v(x,y+1) - v(x,y)) d\Omega
$$
$ l_d $ take the warped first image and second image as input and minimize the difference. $ l_s $ 

\subsection{Moving Object Proposals}
\begin{figure}[ht]
\begin{center}
   \includegraphics[width=1.0\linewidth]{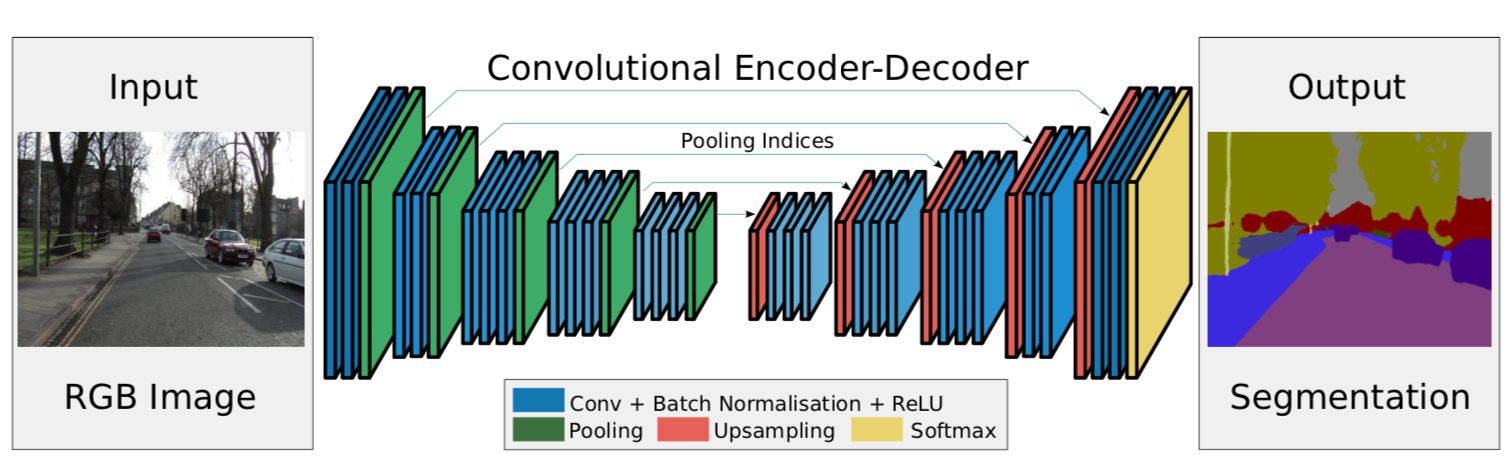}
\end{center}
   \caption{SegNet Architecture}
\label{fig:segnet}
\end{figure}

We propose a temporal stable moving object detector using SegNet architecture. As we can see from Figure~\ref{fig:segnet}, SegNet is separated into two different parts, encoder and decoder. Basically, the encoder uses the first 13 convolutional layers from the VGG16 networks that designed for object classification. And in the decoder part, each layer has a corresponding layer in the encoder, therefore, it also contains 13 layers as the encoder. At the end, the output of decoder is fed to a multi-class soft-max classifier to produce class probability for each pixel. In our project, since our goal is producing segmentation for moving objects rather than semantic segmentation; after getting the input images from the optical flow model, the SegNet model only needs to consider the foreground objects and the background in each image. Therefore, we only need two classes in soft-max function.

In each encoder layer, it performs convolutional with a filter bank, and followed by batch normalization and ReLu, what's more, max-pooling is also used at the end to achieve translation invariance over small spatial shifts in the input image. However, max-pooling causes boundary detail loses. As a solution, the SegNet model stores the max-pooling indices for each encoder layer, and in each decoder layer, it can use corresponding indices for upsampling.

In the training procedure, we train all the variants with Adam Optimizer and the corss-entropy loss function.

\subsection{Expected results and Evaluation} 
We decide to evaluate our results by applying the intersection over union overlap (IoU) between the final results and the ground truth. We can compute IoU according to the formula below,
$$
IoU = A \cap B / A \cup B
$$
where $A$ denotes the area covered on the expected outputs and $B$ denotes the area annotated on ground truth. According to the formula, $IoU$ simply means the proportion of the overlap area within the union area of the model's output and the corresponding ground truth.

\noindent
\textbf{Baselines:} For the evaluation, we decide to compare our method with an advanced methods \textbf{PCM}~\cite{bideau2016s}.

\begin{figure*}
\begin{center}
	\includegraphics[width=1.0\linewidth]{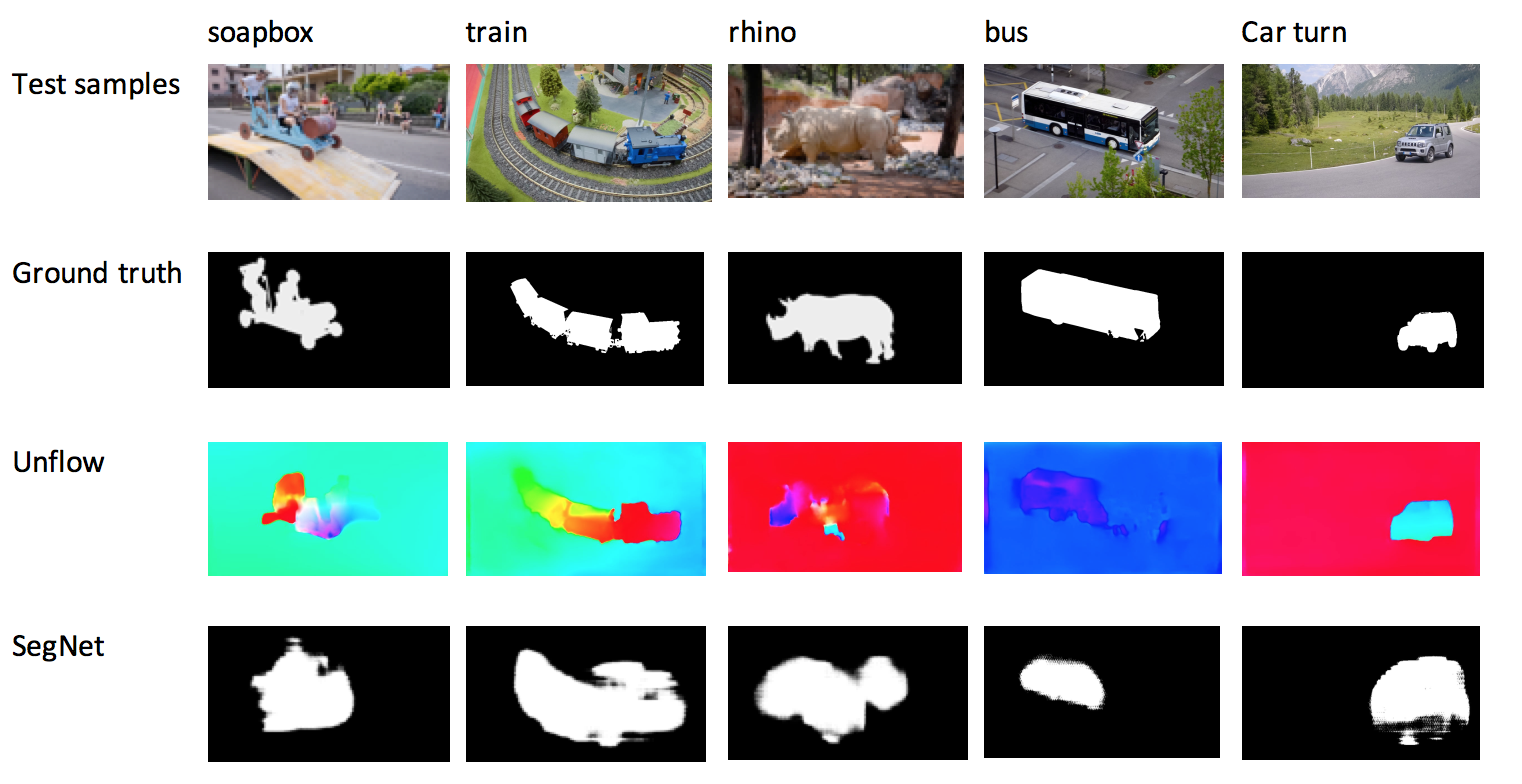}
\end{center}
   \caption{SegNet and Unflow predictions of the samples from DAVIS 2017}
\label{fig:segnet_test}
\end{figure*}
\section{Experiment}
Our model consists of two parts: (1) Unflow net (Unsupervised Learning of Optical Flow); (2) SegNet: A Deep Convolutional Encoder-Decoder Architecture for Image Segmentation. In our implementation, we combined the two models, and modified the source codes for training the dataset DAVIS 2017 \cite{pont20172017}. The evaluation details are as following.

\subsection{Unflow Experiment}
In our implementation, we modified the input and training codes to import the DAVIS 2017 as our training and evaluation dataset. In the training process, only first two images in each classes (50 classes) were chosen to be trained, the training iterations for each class was set as 250.

\subsubsection{Pre-processing}
First, we divide the images in each class into a pair of lists, each pair of images are consecutive frames, therefore, they could be stacked and fed as inputs into the model. What's more, all images need to be reshaped to $448 \times 832$, then applied normalization.
\subsubsection{Early attempts and failures}
\begin{figure}[ht]
\begin{center}
   \includegraphics[width=1.0\linewidth]{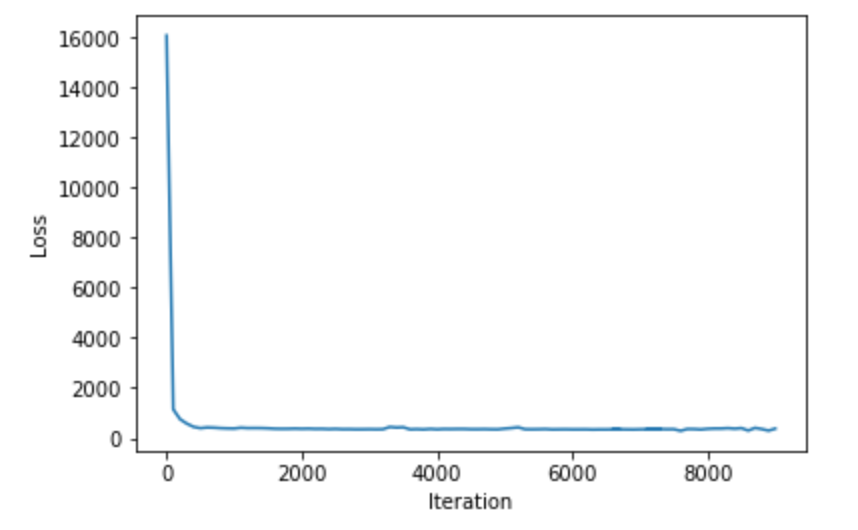}
\end{center}
   \caption{Unflow training loss}
\label{fig:unflowTrainLoss}
\end{figure}

At the beginning, we decided to train our modified model from scratch. The {\it learning rate} was $0.25e^-5$, and the {\it save interval} = 250, so that a checkpoint would be generated in each 250 iterations. Figure \ref{fig:unflowTrainLoss}. displays the losses along with the iterations. As we can see, the training loss dropped dramatically at the beginning, but it stopped reducing at $400$.
In the middle of the training process, we tried to reduce the {\it learning rate} to $0.25e^-6$ and even smaller, but the training loss still could not reduce.
It might be due to the other hyperparameters that take significant roles in the model. But we did not get enough time and resource to have a further test. 

\begin{figure*}
\begin{center}
\includegraphics[width=0.8\linewidth]{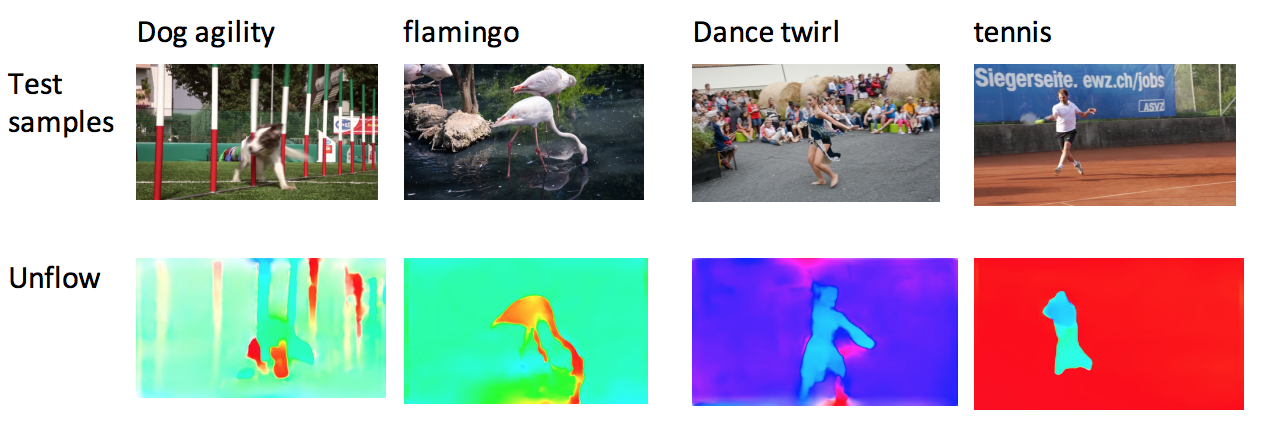}
\includegraphics[width=0.8\linewidth]{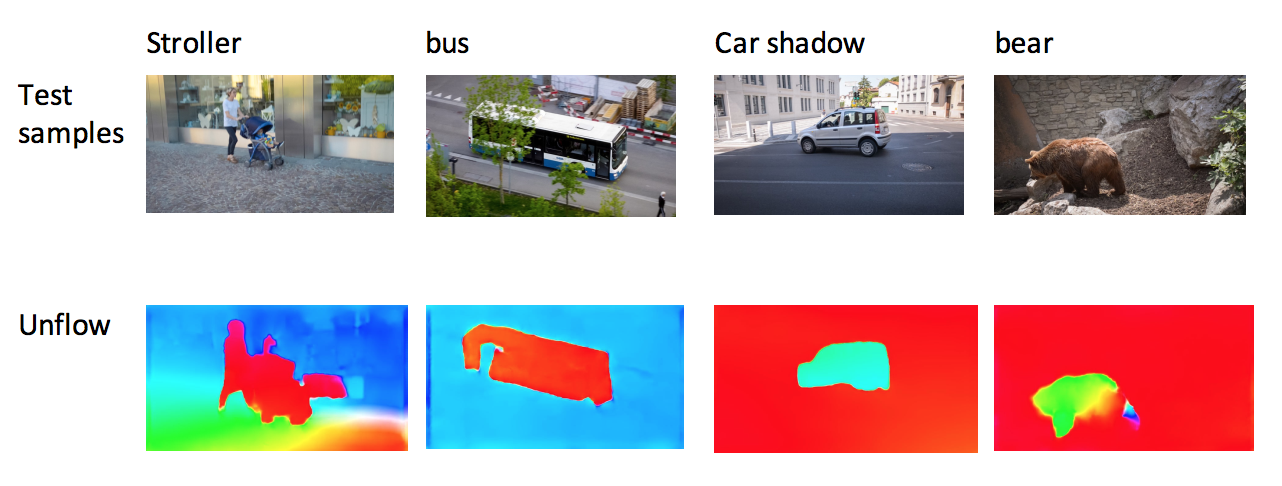}
\end{center}
   \caption{The predicted outputs generated by fine-tuned model of Unflow}
\label{fig:unflow_evaluation}
\end{figure*}
\subsubsection{Successful Approach}
Instead of training everything from scratch, we finally decided to move to a pretrained model that has been fine-tuned. After modifying the codes of evaluation, we successfully adapted our DAVIS 2017 into the pretrained model.
\subsubsection{Results Evaluation}
Although, the pretrained model could provide us with relatively better outputs than our original model, the outputs still consist a lot of glitches that might affect the final results of the whole model. We summarize some common errors occur among those outputs in Figure 7.

\noindent
\textbf{Obstacles:} As we can see from the example \textit{Dog agility} in Figure 7, the unflow outputs includes the pillars as moving objects, which might make sense because while the dog is crossing those pillars, it might also cause the pillars keep vibrating in different frames. As a result, those vibrating pillars are also detected.

\noindent
\textbf{Reflections:} Reflections are another common features in images, for example, in \textit{flamingo}, the reflection of the flamingo on the water is also included in the output.

\noindent
\textbf{Edges:} It is easy for the model to recognize the body of an object, but detecting edge's information always be a big problem for the model. We can find this problem almost in all examples in Figure 7, such as the missing arms and legs in \textit{dance twirl} and \textit{tennis}, the missing wheels in \textit{bus} and \textit{car shadow}.

\subsection{SegNet Experiment}
\subsubsection{Training Details}
We trained the SegNet with Adam Optimizer for 10000 iterations. $ \beta 1 = 0.9 $, $ \beta 2 = 0.999 $, the start learning rate is $ 5e^-4 $, with a decay of 0.5 times every 2000 global steps. We fed the optical flow estimation from Unflow and ground truth segmentation into SegNet and the get the moving object segmentation. 

\begin{figure}[ht]
\begin{center}
   \includegraphics[width=1.0\linewidth]{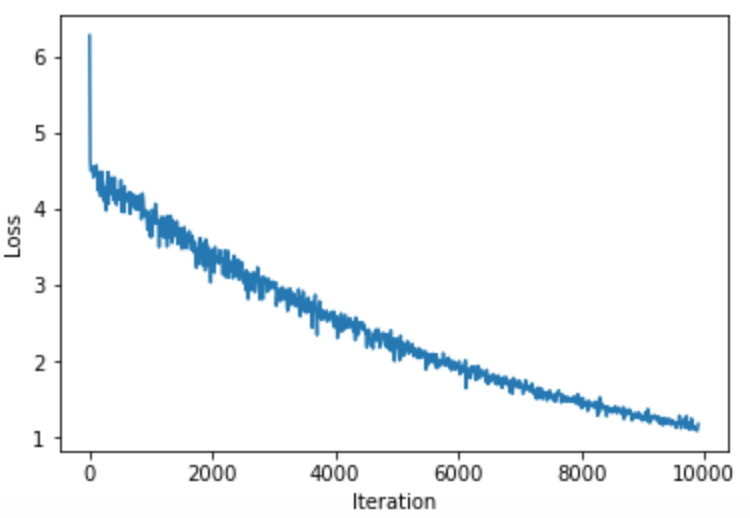}
\end{center}
   \caption{SegNet training loss}
\label{fig:trainLoss}
\end{figure}

\subsubsection{Qualitative Analysis}
We evaluate our results on the evaluation dataset of DAVIS for the lack of test data. The Figure~\ref{fig:trainLoss} shows the loss decreasing with global steps growing. We compared out approach with Table~\ref{table:iou} compares our approach to the state-of-the-art methods on DAVIS. The measure to evaluate results is mean IOU. Because our approach only consider the motion information without semantic information, the result of ours can only be a little better than PCM. When it is put on the test data, we assume it would be worse to some extent.

\begin{table}
\begin{center}
\begin{tabular}{|c|c|c|c|c|c|c|c|c|c|}
\hline
Measure & PCM & CVOS & KEY & NLC & FST & Ours \\
\hline\hline
Mean IOU & 40.1 & 48.2 & 49.8 & 55.1 & 55.8 & 41.9 \\
\hline
\end{tabular}
\end{center}
\caption{Our method compared to other state-of-the-art methods on DAVIS.}
\label{table:iou}
\end{table}

\subsubsection{Quantitative Analysis}
Based on the results of our experiment, we found our networks is more suitable for segment monolithic moving objects with gentle motion. This come from the intrinsic properties of Unflow networks. Also, we found in training process, the segmentation boundaries of moving object tends to shrink with the increasing of training steps. 

\section{Conclusion}

\subsection{Insights}
This paper introduces a novel approach for video moving object segmentation. In general, we tried to combine Unflow and SegNet to get the motion segmentation and the basic idea is working. However, limited by time and resources, we cannot train the networks with a lot more global steps and elaborate parameters. In this work, we leveraged the state-of-the-art unsupervised optical flow estimation approach and fine-tuned it on the novel DAVIS 2017 dataset in which we learned the constrained optical flow formula with smoothness penalty. Besides, we studied the Encoder-Decoder architecture and the advantages of up-sampling and up-convolutional layers.

\subsection{Future Improvement}
The analysis of results inspires us to make improvement in the flowing ways.

\subsubsection{Adding LSTM Model}

Normal convolutional neural networks deal with independent images in an impressive way. However, in sequential video learning tasks, detached CNN cannot preserve the learned features from the previous frames. To solve this problem, ConvLSTM is developed with the motivation to unify CNN and LSTM in one model. ConvLSTM is one kind of LSTM which is specially designed to aid images sequence learning with convolutional layer as input processing cell. The most recent research adopted this model in video object segmentation and outperforms previous work. The convLSTM layer can be inserted between the Encoder and Decoder layer.

\subsubsection{Involving Semantic Information}

Currently, our work feeds the output of Unflow networks to SegNet, but the two network architecture can be simplified by unifying those two. We assume the input image can be rendered into two branches of local network. One is the Encoder part of semantic networks and the other is the dilated convolution layers of Unflow. Than take both tensors into the Decoder architecture. At the loss end, optical flow unsupervised loss and segmentation supervised loss are both considered. In this process, not only the networks can be simplified by removing redundancy but also the semantic information is taken into consider.

\newpage

{\small
\bibliographystyle{ieee}
\bibliography{egbib}
}

\end{document}